\documentclass[10pt,twocolumn,letterpaper]{article}

\usepackage{iccv}
\usepackage{times}
\usepackage{epsfig}
\usepackage{graphicx}
\usepackage{amsmath}
\usepackage{amssymb}

\usepackage{microtype}
\usepackage{graphicx}
\usepackage{subfigure}
\usepackage{booktabs} 
\usepackage{multirow}

\usepackage{amsmath,amssymb}
\usepackage[dvipsnames]{xcolor}
\usepackage{eso-pic}

\usepackage[breaklinks=true,bookmarks=false]{hyperref}

\iccvfinalcopy 

\def\iccvPaperID{****} 

\begin{document}

\title{Efficient Decoupled Neural Architecture Search \\ by Structure and Operation Sampling}

\author{Heung-Chang Lee\thanks{Equal contribution} $^\dagger$~~ Do-Guk Kim$^\ast$\thanks{Corresponding authors}
\\ BigData \& AI Lab \\ Hana Institute of Technology, Hana TI, Korea \\
{\tt\small leehc.com@gmail.com, logue311@gmail.com}
\and
Bohyung Han\\
Dept. of Electrical and Computer Engineering \\ Seoul National University, Korea \\
{\tt\small bhhan@snu.ac.kr}
}

\maketitle

\begin{abstract}
We propose a novel neural architecture search algorithm via reinforcement learning by decoupling structure and operation search processes.
Our approach samples candidate models from the multinomial distribution on the policy vectors defined on the two search spaces independently.
The proposed technique improves the efficiency of architecture search process significantly compared to the conventional methods based on reinforcement learning with the RNN controllers while achieving competitive accuracy and model size in target tasks.
Our policy vectors are easily interpretable throughout the training procedure, which allows to analyze the search progress and the discovered architectures; the black-box characteristics of the RNN controllers hamper understanding training progress in terms of policy parameter updates.
Our experiments demonstrate outstanding performance compared to the state-of-the-art methods with a fraction of search cost.
\end{abstract}

\section{Introduction}
\label{sec:introduction}
Designing deep neural network architectures often requires various task-specific domain knowledge, and it is challenging to achieve the state-of-the-art performance by manual tuning without such information.
Consequently, the automatic search for network architectures becomes an active research problem~\cite{stanley2002evolving,andrychowicz2016learning,li2016learning}.
Several neural architecture search (NAS) techniques achieve the state-of-the-art performances on the standard benchmark datasets~\cite{zoph2018learning,chen2018searching}.
However, NAS methods inherently suffer from high computational cost due to their huge search spaces for architectural variations and frequent performance evaluation requirement during training.
To overcome these limitations, weight sharing concept has recently been proposed and illustrated its advantage in terms of accuracy and efficiency~\cite{pham2018efficient}.
Despite such efforts, the computational cost of NAS approaches is still too prohibitive to be applied to the search problems for large-scale models and/or datasets.
Another critical drawback of the most existing methods is that it is extremely difficult to understand their training progress since decision making typically relies on the hidden state representations of the architecture search models.

We propose an efficient decoupled neural architecture search (EDNAS) algorithm based on reinforcement learning (RL).
Contrary to the conventional RL-based NAS methods, which employ an RNN controller to sample candidate architectures from the search space, we use the policy vectors for decoupled sampling from the structure and operation search spaces.
The decoupled sampling strategy enables us to reduce search cost significantly and analyze the architecture search procedure in a straightforward way.
The resulting architecture achieves competitive performance compared to the output models from the state-of-the-art NAS techniques.

We claim the following contributions in this paper: 
\begin{itemize}
    \item We propose an RL-based neural architecture search technique, which learns the policy vectors to samples candidate models by decoupling structure and operation search spaces of a network.
    \item Our sampling strategy is based on the fully observable policy vectors over the two orthogonal search spaces, which facilitates to analyze the architecture search progress and understand the learned models.
    \item Our algorithm achieves the competitive performances on various benchmark datasets including CIFAR-10, ImageNet, and Penn Treebank with a fraction of computational cost.
\end{itemize}

The rest of this paper is organized as follows.
In Section~\ref{sec:related}, we discuss the related work.
We describe the proposed algorithm in Section~\ref{sec:methodology}, and then illustrate experimental results in Section~\ref{sec:experiments}.
Section~\ref{sec:conclusion} concludes our paper.

\section{Related Work}
\label{sec:related}
Existing NAS methods are categorized into three groups based on their search methodologies: RL-based  ~\cite{zoph2016neural,zoph2018learning,pham2018efficient,zhong2018practical,liu2018progressive, perez2018efficient}, evolutionary algorithm (EA)-based  ~\cite{real2017large,real2018regularized,elsken2018efficient,so2019theevolved}, and gradient-based  ~\cite{liu2018darts,luo2018neural,zhang2018you,xie2018snas,cai2018proxylessnas}.
We summarize the technical details of each approach.

\subsection{RL-based Methods}
RL-based architecture search techniques are originally proposed in ~\cite{zoph2016neural}, where the RNN controller is used to search for whole models and the feedback from validation is given by reinforcement learning.
NASNet~\cite{zoph2018learning} follows the optimization framework of ~\cite{zoph2016neural}, but the construction of the network is based on the cells discovered by the RNN controller.
The search space reduction to a cell is proven to improve not only efficiency but also accuracy compared to the unrestricted exploration ~\cite{zoph2018learning,bender2018understanding,pham2018efficient}.
However, NASNet still demand a lot of search cost, which makes the algorithm impractical.
ENAS~\cite{pham2018efficient} aims to further improve efficiency by weight sharing. 
It achieves a remarkable reduction in search cost compared to NASNet with competitive accuracy.

\subsection{EA-based Methods}
The most representative work of EA-based approach is ~\cite{real2017large}, where a CNN is evolved from the trivial simple architecture resulting in comparable image classification performance to the state-of-the-art models.
This technique is extended to evolving the convolutional cells rather than the whole architecture, which is referred to as AmoebaNet~\cite{real2018regularized}.
The best discovered architecture of AmoebaNet achieves the outstanding performance on ImageNet.
Recently, the evolution from the known architecture has been investigated to achieve the improved accuracy in ~\cite{so2019theevolved}.

\subsection{Gradient-based Methods}
The gradient-based approach is an emerging direction in the NAS field, and many recent works can be classified into this category.
These methods relax a discrete architecture space into a continuous one and apply a gradient-based optimization technique to find the best model.
The relaxation is achieved by constructing a network structure based on various operations with differentiable weights~\cite{liu2018darts}, encoding network architectures using feature vectors in a continuous space~\cite{luo2018neural}, and adopting the concrete distribution over network architectures~\cite{xie2018snas}.

\subsection{Others}
There are other types of methods, which include hypernetworks~\cite{brock2017smash, zhang2018graph} and efficient random search techniques~\cite{bender2018understanding}.

\section{Methodology}
\label{sec:methodology}
EDNAS is an RL-based NAS approach with reduced complexity and observable search mechanism while maintaining accuracy in target tasks.
We first define our search space in Section~\ref{sect2.1}, and then discuss the details of the policy vectors and our search algorithm in Section~\ref{sect2.2}.
The training procedure is described in Section~\ref{sect2.3}.

\begin{figure}[t]
\begin{center}
\includegraphics[width=0.8\linewidth]{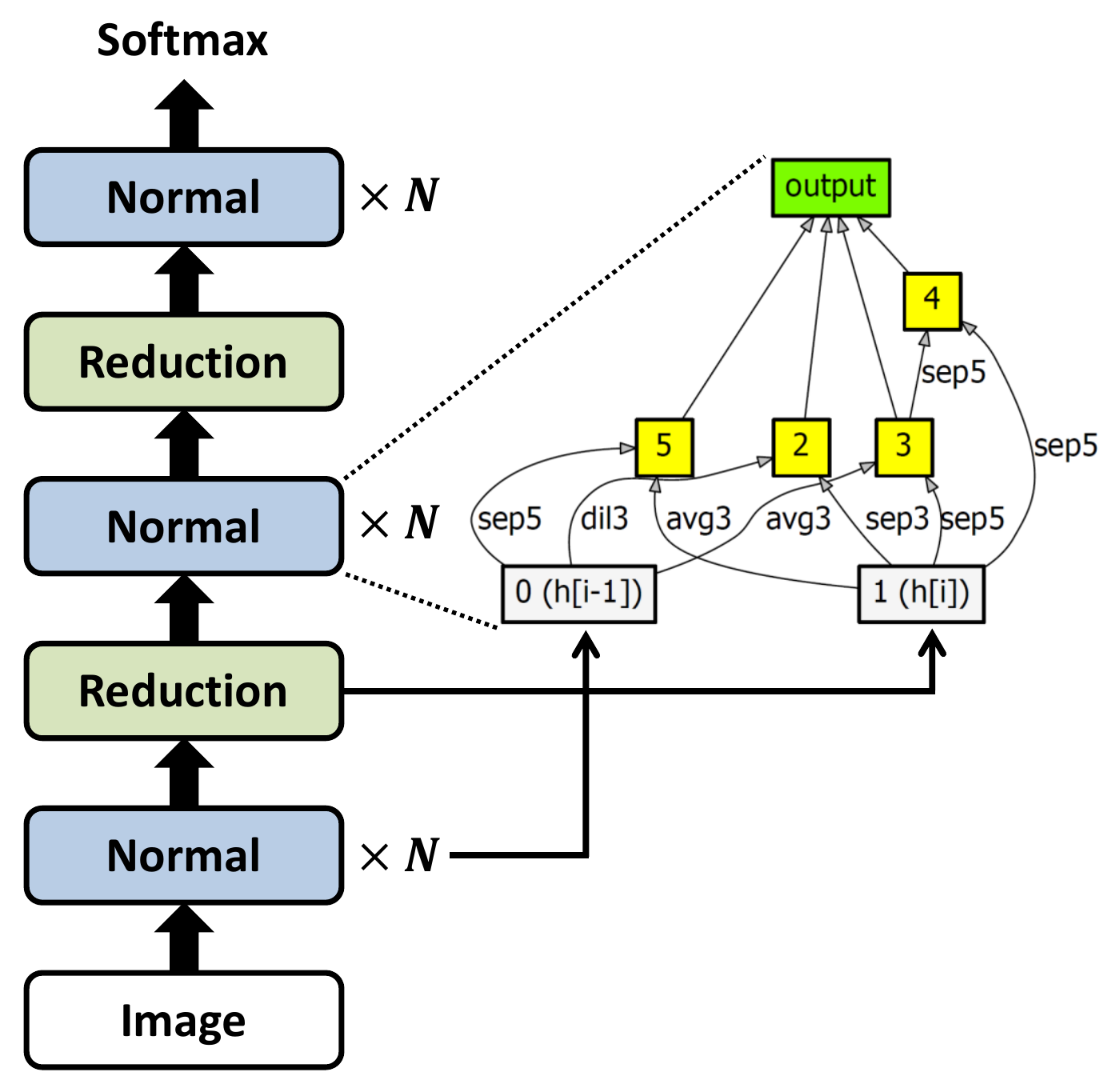}
\vspace{0.1cm}
\caption{The overall architecture of the convolutional network.}
\label{overall_and_DAG}
\end{center}
\end{figure}

\subsection{Overall Architecture Design}

According to recent studies, it is inefficient to search the entire architecture of neural network~\cite{zoph2018learning, pham2018efficient, real2018regularized}.
This search strategy is called a macro search method~\cite{pham2018efficient}.
In opposite, micro search means that search only cell architectures and stack discovered cells to construct the entire architecture.
In several previous works~\cite{zoph2018learning, pham2018efficient}, it is proven that micro search has better final performance than macro search even its search cost is lower.
Therefore, we adopted the micro search strategy in EDNAS.

Our overall architecture design is presented in Fig~\ref{overall_and_DAG}.
As presented in Fig~\ref{overall_and_DAG}, the entire convolutional network is consists of stacked cells.
There are two sorts of cells which are normal cell and reduction cell.
A normal cell has the same input and output size, and reduction cell has an output size of half of the input width and height.
By changing the repeating number of normal cells, we can change the parameter size of the overall network.
In the case of the recurrent cell, we used a single cell for the entire recurrent network.

\subsection{Search Space}
\label{sect2.1}
The search space of EDNAS is given by a cell structure as in the recent studies~\cite{zoph2018learning, pham2018efficient, liu2018darts}.
The cell architecture is defined by a directed acyclic graph (DAG), where the nodes and the edges represent the local computation results and the flow of activations, respectively.
The search process starts with a manually designed overall architecture, which is a stack of cells in case of convolutional neural networks, and a sequence of cells in recurrent networks.

A convolutional cell has two input nodes, which have no incoming edges in the DAG, and they correspond to the outputs of two previous cells.
A node takes the outputs of two previous nodes as its inputs, and applies operations to individual inputs.
Then, inputs are summed to generate the output of the node.
The input-output relations are defined by the edges in the DAG.
We consider seven operations as in the existing methods~\cite{liu2018darts, pham2018efficient, luo2018neural}, which include $3\times3$ and $5\times5$ separable convolutions, $3\times3$ and $5\times5$ dilated separable convolutions, a $3\times3$ max pooling, a $3\times3$ average pooling, and an identity connection.
The output of the convolutional cell is defined as the depth-wise concatenation of all hidden nodes.

A recurrent cell also has two input nodes, which correspond to the input of the current cell and the hidden state of the previous cell.
Each node takes an output of the preceding one, performs an operation given by an edge, and then produces its own output after applying the activation function.
We employ four activation functions commonly used in previous studies~\cite{liu2018darts, pham2018efficient, luo2018neural}: sigmoid, tanh, ReLU, and identity.
To construct a recurrent cell, we use eleven nodes with nine hidden nodes.
The output of the recurrent cell is defined as the average of all hidden nodes.

\begin{figure*}[t!]
\begin{center}
\centerline{\includegraphics[width=0.83\linewidth]{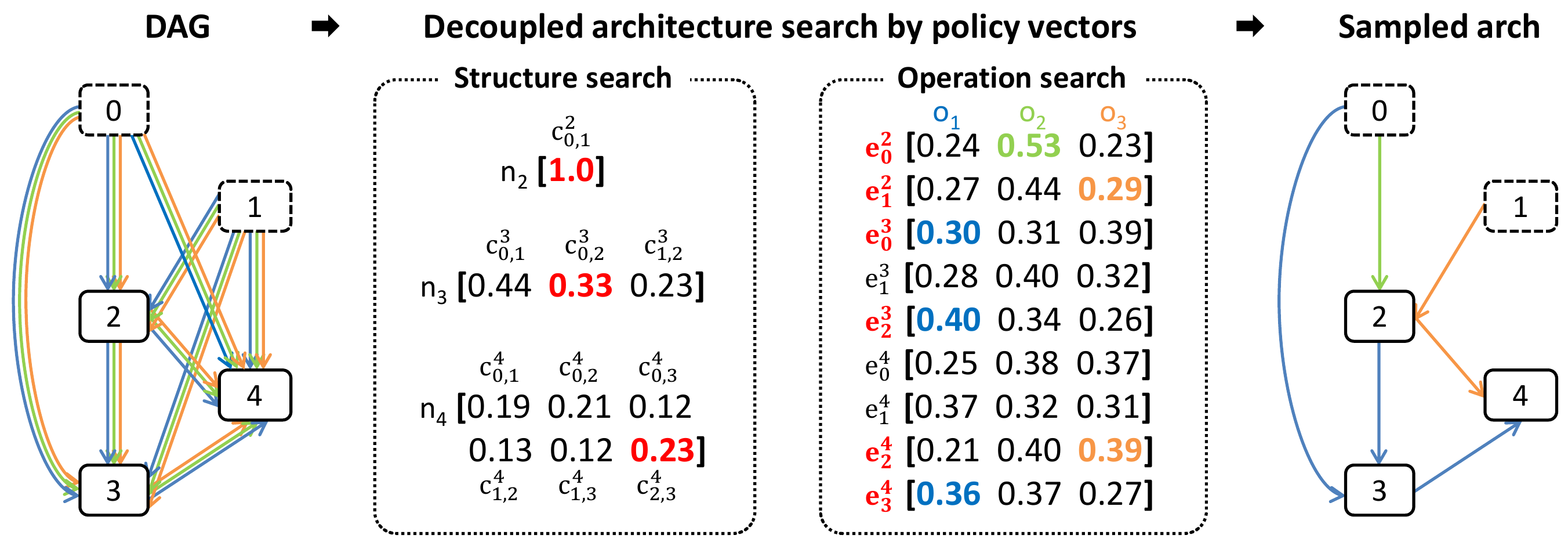}}
\caption{An example of the convolutional cell sampling by EDNAS when $N=5$. Input nodes are represented with the dashed rectangles while non-input nodes are represented with the solid rectangles. From each policy vector with softmax, the actions are sampled by multinomial sampling. The selected edges and operations are highlighted with colors in the policy vectors.}
\vspace{-0.5cm}
\label{cnn_sampling}
\end{center}
\end{figure*}

\begin{figure*}[t!]
\begin{center}
\centerline{\includegraphics[width=0.83\linewidth]{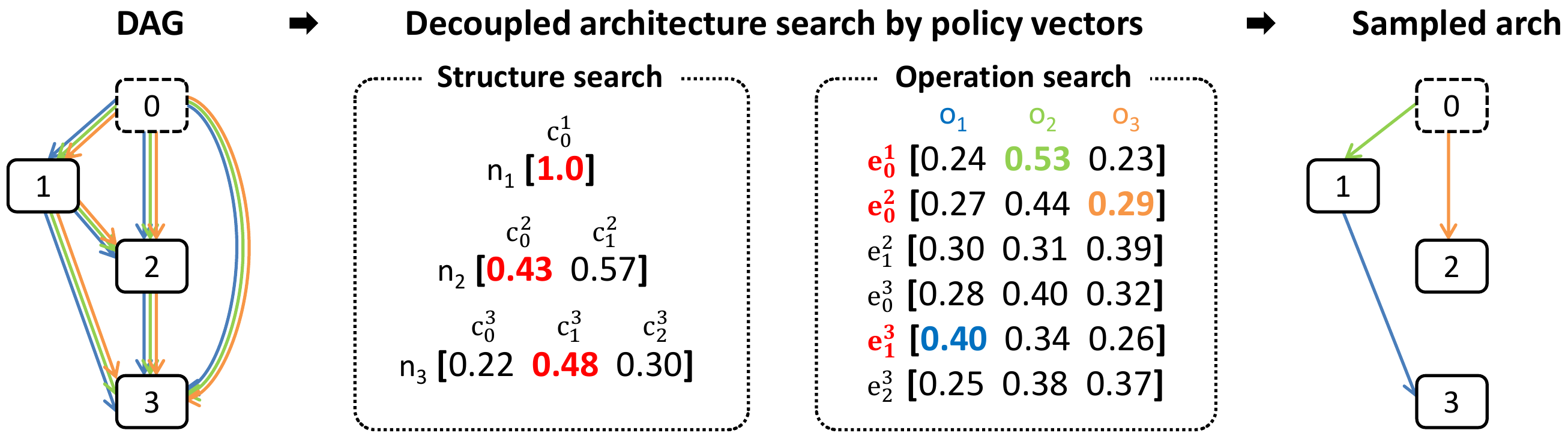}}
\caption{An example of the recurrent cell sampling by EDNAS when $N=4$. The first hidden node is represented with the dashed rectangle, and other hidden nodes are represented with the solid rectangles. From each policy vector with softmax, the actions are sampled by multinomial sampling. The selected edges and operations are highlighted with colors in the policy vectors.}
\vspace{-0.5cm}
\label{rnn_sampling}
\end{center}
\end{figure*}

\subsection{Architecture Search Process in EDNAS}
\label{sect2.2}

\subsubsection{Overall Process}
The objective of our algorithm, EDNAS, is to maximize the expected reward of the sampled architectures~\cite{zoph2016neural}, which is given by
\begin{equation}
\label{EDNAS_objective}
   \max_{\theta}~ \mathbb{E}_{P(\mathbf{m};\theta)}[R],
\end{equation}
where $\theta$ is the policy parameter and $\mathbf{m}$ is the sample model based on the current policy.
While the RNN controller manages policies in the conventional RL-based methods ($\theta = \theta_c$), EDNAS employs the policy vectors instead ($\theta = \theta_v$) to search for the optimal architecture. 

EDNAS decouples the structure search and operation search, which are performed based on the separate policy vectors.
There are two kinds of policy vectors in EDNAS; one is for non-input nodes and the other is for edges.
All non-input nodes in the DAG are associated with a $c_i$-dimensional policy vector, where $c_i$ is the number of incoming edge combinations to the $i$-th node.
The policy vector of node $n_i$, $\mathbf{p}_{n_i}$, is given by
\begin{equation}
\label{node_policies}
    \mathbf{p}_{n_i}=[v_1, ..., v_{c_i}],  \;\; c_i={e_i\choose r},
\end{equation}
where $e_i$ is the number of incoming edges to $n_i$, and $r$ is the number of the selected edges. 
The policy vector of edge $e$ in the DAG is a $k$-dimensional vector, where $k$ is the number of operations:
\begin{equation}
\label{edge_policies}
    \mathbf{p}_{e}=[w_1, ..., w_{k}].
\end{equation}

Based on these policy vectors, we perform architecture sampling as follows.
First, we search for the overall structure of the network by sampling edges from the entire DAG.
To this end, the softmax function is applied to $\mathbf{p}_{n_i}$ for its normalization.
Then, an input edge combination of each node is sampled from the multinomial distribution given by $\text{softmax}(\mathbf{p}_{n_i})$.
After that, we optimize the operation corresponding to each selected edge.
The operation of each selected edge is determined by drawing a sample from the multinomial distribution defined by $\text{softmax}(\mathbf{p}_{e})$, which is similar to the structure searching step.

The policy vectors for the structure and operation search are observable in our framework.
Therefore, we can analyze the training progress of EDNAS based on the statistics of architecture samples and the visualized policy vectors during training.
For example, it is possible to see which combinations of edges are selected and which operations are preferred at each iteration or over many epochs.

\subsubsection{Searching in Convolutional Cells}
Figure~\ref{cnn_sampling} illustrates an example of convolutional cell architecture sampling.
The number of nodes \textit{N} is 5, and the number of operations is 3.
There are two input nodes in the DAG as described in Section~\ref{sect2.1}, and non-input nodes in our convolutional cell receive two inputs from preceding nodes, {\it i.e.,} $r=2$.
In the structure search step, one input edge combination is selected for each non-input node by multinomial sampling.
In the example, the edges heading to $n_3$ from $n_0$ and $n_2$ are selected as the input edge combination of $n_3$.
For the node $n_4$, edges from $n_2$ and $n_3$ are chosen.

In operation search step, the specific operation of the selected edges is determined.
Note that we are supposed to select an operation for each selected edge.
We obtain a sampled architecture after both steps are completed, as shown in the rightmost graph in Figure~\ref{cnn_sampling}.
Since the recent NAS methods typically adopt both normal and reduction cell structure to construct convolutional networks, we also search for both cells based on the policy vectors defined separately.

The computational complexity of the architecture search in convolutional cells is estimated by the number of candidate architectures.
In case of EDNAS, a total number of possible edge combinations is $\prod_{i=1}^{N-2} {i+1\choose 2}$ and a number of cases of operation combinations in an edge combination is $7^8$.
Therefore, the number of possible convolutional cell architecture of EDNAS is $\prod_{i=1}^{N-2} {i+1\choose 2} \cdot 7^8 = 1.04 \times 10^9$ since we use $N=6$.

\subsubsection{Searching in Recurrent Cells}
Sampling procedure in the recurrent cell architectures is presented in Figure~\ref{rnn_sampling}.
In the example, the number of nodes $N$ is 4 and the number of activation functions is 3.
Each non-input nodes takes an incoming edge from one of the preceding nodes ($r=1$).
The first hidden node, denoted by $n_0$ in Fig.~\ref{rnn_sampling}, adds up two input nodes and applies the $\text{tanh}$ activation to compute its output.

Similarly to the convolutional cell architecture sampling, the edges and activation functions are selected by multinomial distribution based on the policy vectors after applying the softmax function.
In case of the example in Figure~\ref{rnn_sampling}, the edge entering $n_2$ from $n_0$ and the edge heading to $n_3$ from $n_1$ are selected in the structure searching step.
After the operation searching step, a sampled architecture is determined as shown in the graph on the right of Figure~\ref{rnn_sampling}.
The computational complexity is also given by the number of possible recurrent cell architectures of EDNAS, which is $\prod_{i=1}^{N-1} {i\choose 1} \cdot 4^8 = 2.64 \times 10^9$ since we use $N=9$.

\subsection{Training Process and Deriving Architectures}
\label{sect2.3}
In EDNAS, the entire training process consists of the child model training step and the policy vector training step.
We perform the optimization by alternating the two training steps, where the child model sampled from the DAG is learned using the training data while the policy vectors are trained with the validation set.

For training the child model, we use stochastic gradient descent (SGD) on the shared parameters of the DAG to minimize the expected loss.
During the child model training step, the parameters of the policy vectors are fixed.
As mentioned in ENAS~\cite{pham2018efficient}, the gradient given by the Monte Carlo estimate based on a single sample works fine to train the child model.
Therefore, we sample one architecture in every iteration, compute a gradient based on the sampled model, and train the model using SGD.

In the policy vector training step, the model parameters are fixed and the policy vectors are updated to maximize the expected reward.
We use Adam optimizer~\cite{kingma2014adam}, and the gradient is computed by REINFORCE~\cite{williams1992simple}, as shown below:
\begin{equation}
\label{reinforce}
    \nabla_{\theta_v}\log P(\mathbf{m};\theta_v)(R-b),
\end{equation}
where $P(\mathbf{m};\theta_v)$ is the probability of the model $\mathbf{m}$ sampled based on the policy vectors $\theta_v$, and $b$ is a moving average baseline of rewards.
We calculate the reward $R$ on the validation set, and encourage policy vectors to learn architecture with high generalization performance.
In the case of the image classification experiment, we use validation accuracy on a single minibatch as a reward.
In the language modeling, we employ $c \cdot \left( \text{ppl}_\text{valid} \right)^{-1}$  as a reward, where $c$ is a pre-defined constant, and $\text{ppl}_\text{valid}$ is the perplexity on a single validation minibatch.

The training process is repeated until the training epoch reaches the pre-defined maximum number of epochs.
When deriving the final architecture, we sample a predefined number of models and compute rewards of each model on a single minibatch.
Then, the model that achieves the best reward is selected as the final architecture discovered by EDNAS.
We employ 100 models to obtain the final model in both convolutional and recurrent cell search.

\begin{table*}[t]
\caption{Comparison between EDNAS and the state-of-the-art neural architecture search methods for image classification on CIFAR-10 dataset with respect to computational cost, params and test errors. Additionally, $\dagger$ marks are the performance we conducted experiments again in our environment by using codes which made by the author from their GitHub.}
\label{cifar_result}
\vskip 0.15in
\begin{center}
\scalebox{0.85}{
\begin{tabular}{@{}clcccc@{}}
\toprule
& \multicolumn{1}{c}{}& \multicolumn{1}{c}{Search Cost} & \multicolumn{1}{c}{Params} & \multicolumn{1}{c}{Test Error} & \multicolumn{1}{c}{}\\
Category & \multicolumn{1}{l}{Method} & \multicolumn{1}{c}{(GPU days)} & \multicolumn{1}{c}{(M)}& \multicolumn{1}{c}{(\%)} & Search Method\\
\midrule
\multirow{2}{*}{Manual} & DenseNet~\cite{huang2017densely} & - & 25.6 & 3.46 & manual \\
&DenseNet + cutout~\cite{huang2017densely} & - & 26.2 & 2.56 & manual \\
\midrule
\multirow{5}{*}{\shortstack{NAS without \\ weight sharing}} & NASNet-A + cutout~\cite{zoph2018learning} & 1800 & 3.3 & 2.65 & RL\\
&AmoebaNet-A + cutout~\cite{real2018regularized} & 3150 & 3.2 & 3.34 & EA\\
&AmoebaNet-B + cutout~\cite{real2018regularized} & 3150 & 3.1 & 2.55 & EA\\
&PNAS~\cite{liu2018progressive} & 225 & 3.1 & 3.41 & RL\\
&NAONet + cutout~\cite{luo2018neural} & 200 & 128 & 2.11 & gradient\\
\midrule
\midrule
\multirow{10}{*}{\shortstack{NAS with \\weight sharing}} & ENAS + cutout~\cite{pham2018efficient} & 0.5 & 4.6 & 2.89 & RL\\
&ENAS + cutout$^\dagger$~\cite{pham2018efficient} & 0.6 & 3.2 & 3.32  & RL\\
&DARTS (first order) + cutout~\cite{liu2018darts} & 0.38 & 2.9 & 2.94 & gradient \\
&DARTS (first order) + cutout$^\dagger$~\cite{liu2018darts} & 0.32 & 2.8 & 3.05 & gradient \\
&DARTS (second order) + cutout~\cite{liu2018darts} & 1 & 3.4 & 2.83 & gradient\\
&NAONet-WS~\cite{luo2018neural} & 0.3 & 2.5 & 3.53 & gradient\\
&GHN + cutout~\cite{zhang2018graph} & 0.84 & 5.7 & 2.84 & hypernet \\
&DSO-NAS + cutout~\cite{zhang2018you} & 1 & 3.0 & 2.84 & gradient \\
\cmidrule{2-6}
&Random & 0.27 & 3.4 & 3.91 & -\\
&EDNAS + cutout & 0.28 & 3.7 & 2.84 & RL\\
\bottomrule
\end{tabular}
}
\end{center}
\vskip -0.1in
\end{table*}

\subsection{Characteristics of EDNAS}
EDNAS has a more interpretable architecture search procedure compared to other RL-based methods searching for architectures using RNN controllers~\cite{zoph2018learning,pham2018efficient}. 
This is because our algorithm maintains two decoupled policy vectors---one for structure search and the other for operation search---to sample architectures; these policy vectors are human interpretable and the search procedures are fully observable.

Another benefit of decoupling the policy vectors is the reduction of architecture search cost by the projection of search spaces.
Note that the methods relying on RNN controllers~\cite{zoph2018learning,pham2018efficient} need to consider all the generated architecture sequences while a gradient-based method~\cite{liu2018darts} consider all possible combinations of architectures during training and has to construct a huge model for neural architecture search. 
In practice, the running time of EDNAS is faster than other methods, and the gaps become larger in the large-scale network architecture search.

\begin{figure*}[t!]
\begin{center}
\centerline{\includegraphics[width=\linewidth]{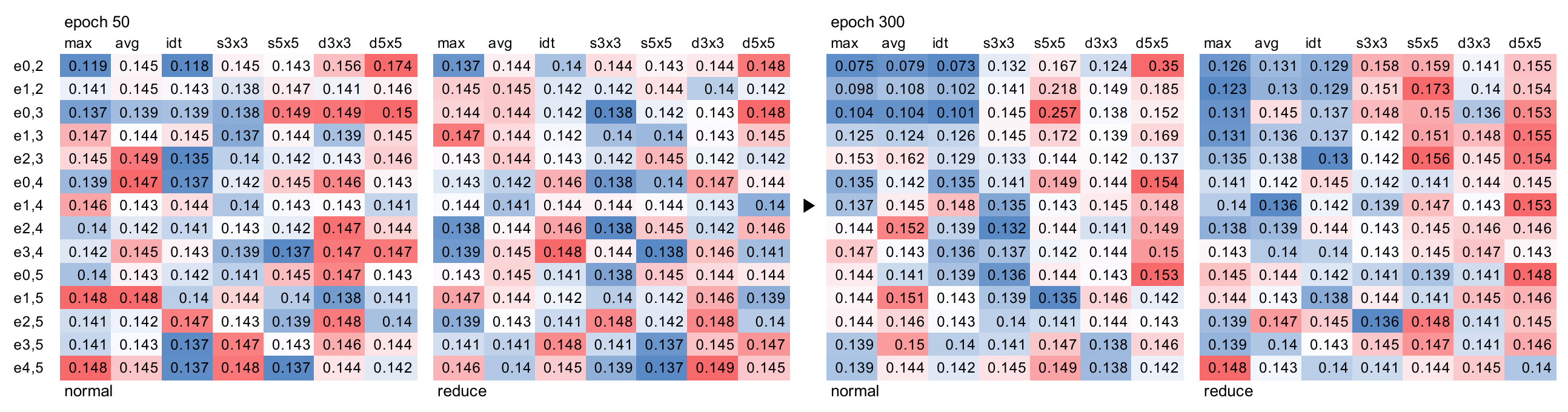}}
\vspace{-0.1cm}
\caption{An example of policy vectors for sampling operations in a time sequence of epochs on CIFAR-10 dataset. This table defines columns as operations and rows as connecting to each edge.}
\vspace{-0.5cm}
\label{cifar_heatmap}
\end{center}
\end{figure*}

\section{Experiments}
\label{sec:experiments}
We conducted experiments on CIFAR-10 and ImageNet dataset to identify the optimal convolutional models, and on Penn Treebank (PTB) dataset to search for recurrent networks.
Refer to our project page\footnote{https://github.com/logue311/EDNAS} for the source code which is able to facilitate the reproduction of our results.

\subsection{Convolutional Cell Search with CIFAR-10}

\subsubsection{Data and Experiment Setting} 
The CIFAR-10 dataset has 50,000 training images and 10,000 testing examples. 
Among the training examples, EDNAS uses 40,000 images for training models and 10,000 images for training policy vectors in the architecture search.

In architecture search, EDNAS uses 2 input nodes and 4 operation nodes to design the architecture in a cell.
We construct the whole architecture using 6 normal cells and 2 reduction cells and each reduction cell is located after 2 normal cells.
%
Our approach utilizes the following hyper-parameters.
For training child models, we employ SGD with the Nesterov momentum~\cite{nesterov1983} with the learning rate 0.05 and the batch-size 128.
For learning the policy vectors, the optimization is given by the Adam~\cite{kingma2014adam} with the learning rate 0.00035.
Both child models and the policy vectors are trained for 300 epochs.

\begin{figure*}[t]
  \centering
  \subfigure[The edge sampling statistics of the normal cells.]{
  \includegraphics[width=0.42\linewidth]{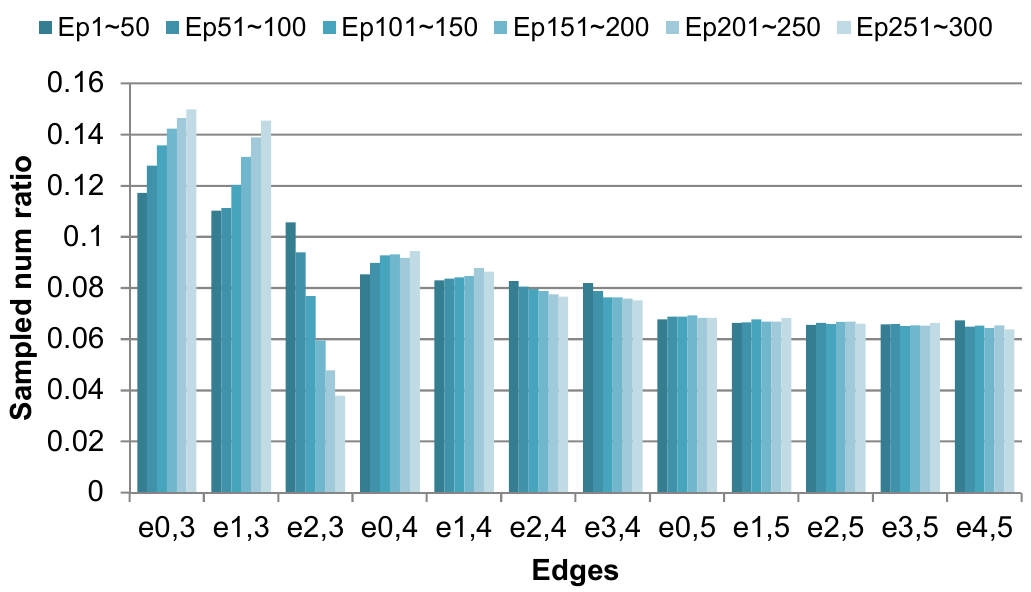}
  \label{edges_stat_normal}}
  \hspace{0.4in}
  \subfigure[The edge sampling statistics of the reduction cells.]{
  \includegraphics[width=0.42\linewidth]{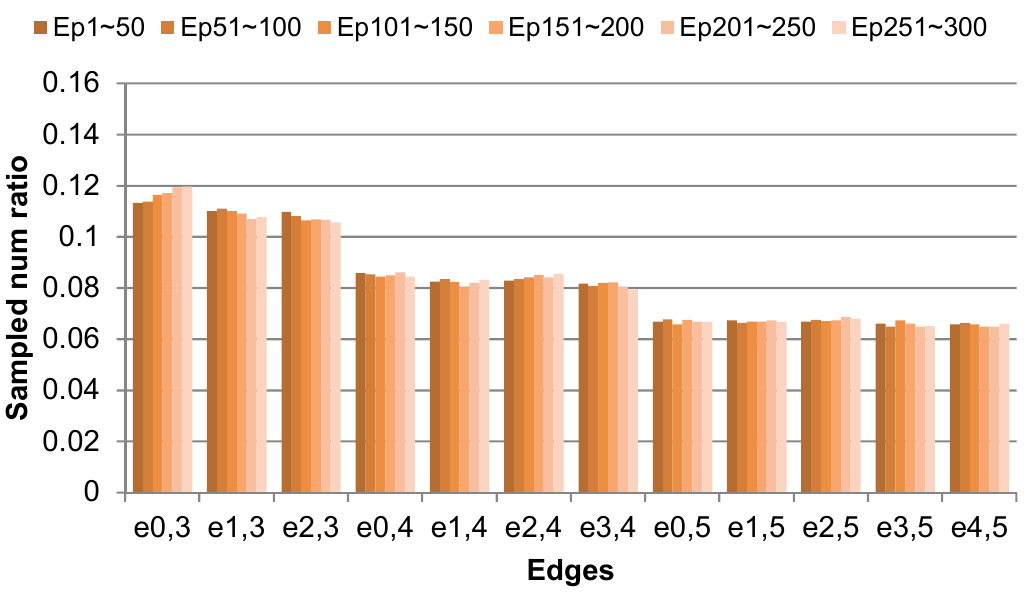}
  \label{edges_stat_reduction}}
  \subfigure[The operation sampling statistics of the normal cells.]{
  \includegraphics[width=0.42\linewidth]{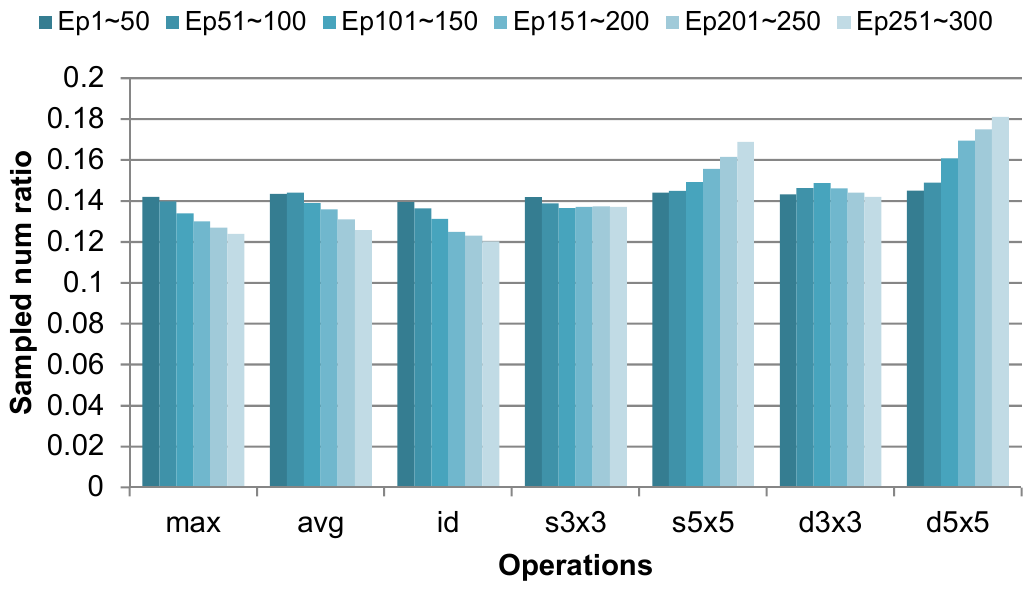}\label{operations_stat_normal}}
  \hspace{0.4in}
  \subfigure[The operation sampling statistics of the reduction cells.]{
  \includegraphics[width=0.42\linewidth]{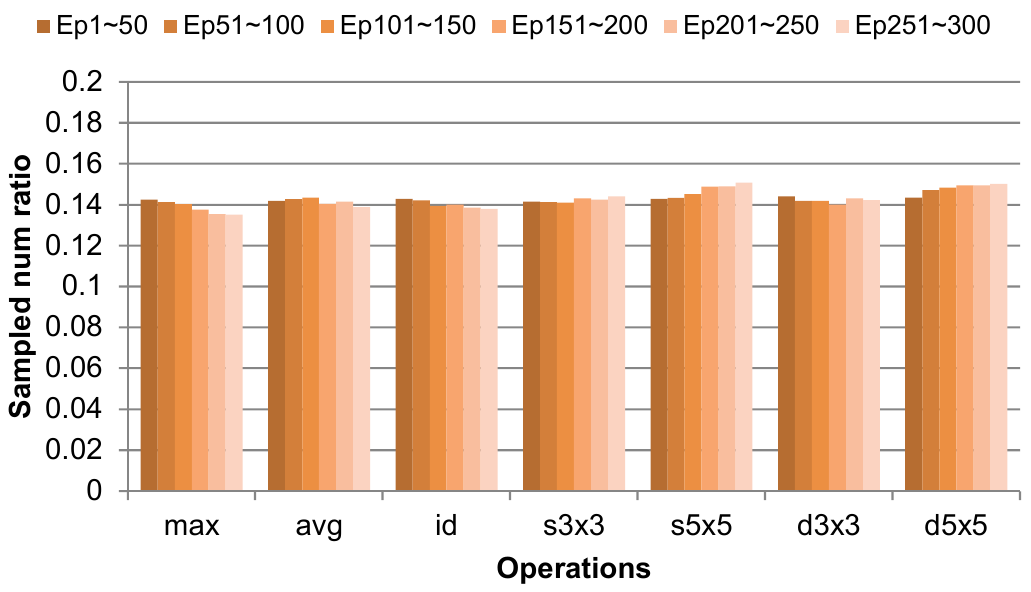}
  \vspace{-0.1cm}
  \label{operations_stat_reduction}}
  \caption{The cumulative histograms of the sampled edges and operations for every 50 epoch. The sampling results in the normal cells show the clear tendency while the sampling results in the reduction cells are mostly stable over time.}
  \label{histograms}
\end{figure*}

We adopt a different architecture for performance evaluation, which is composed of 20 cells (18 normal cells and 2 reduction cells); 6 normal cells are followed by a reduction cell twice, and there are 6 normal cells with the auxiliary classifier at the end of the network to reduce the vanishing gradient problem.
The learning rate is 0.025, the batch size is 128, and the network is trained for 600 epochs.
We optimize the network using SGD without the Nesterov momentum and incorporate the cutout~\cite{devries2017improved} method for better generalization. 

\subsubsection{Results and Discussion}
Table~\ref{cifar_result} summarizes the results.
Although the manual search algorithms achieve the state-of-the-art accuracy, the model sizes are much larger than automatic architecture search techniques.
The NAS methods without weight sharing suffer from huge search cost while the approaches with weight sharing have a reasonable balance between cost and accuracy.
Note that EDNAS achieves competitive accuracy to the techniques with weight sharing in terms of accuracy and model size, but it is substantially faster for architecture search.

Our architecture search procedures are fully observable and Figure~\ref{cifar_heatmap} visualizes the policy vector for sampling operations during training on CIFAR-10 dataset.
The policy vectors are represented as a matrix, where each column denotes an operation and each row means an edge in the DAG. 
Each edge is identified by its source and destination, which are shown as the numbers in the row label.
The number in each cell is the value after applying the softmax function.
The background color and its transparency of each cell are normalized within the cells of the same destination nodes.
The dark red means a large number while the dark blue is a small one.
Note that the distribution of values in the policy vectors are almost random at epoch 50 while the policy vectors at epoch 300 prefer convolution operations than non-trainable operations.

\begin{figure}[t]
\begin{center}
\includegraphics[width=0.6\linewidth]{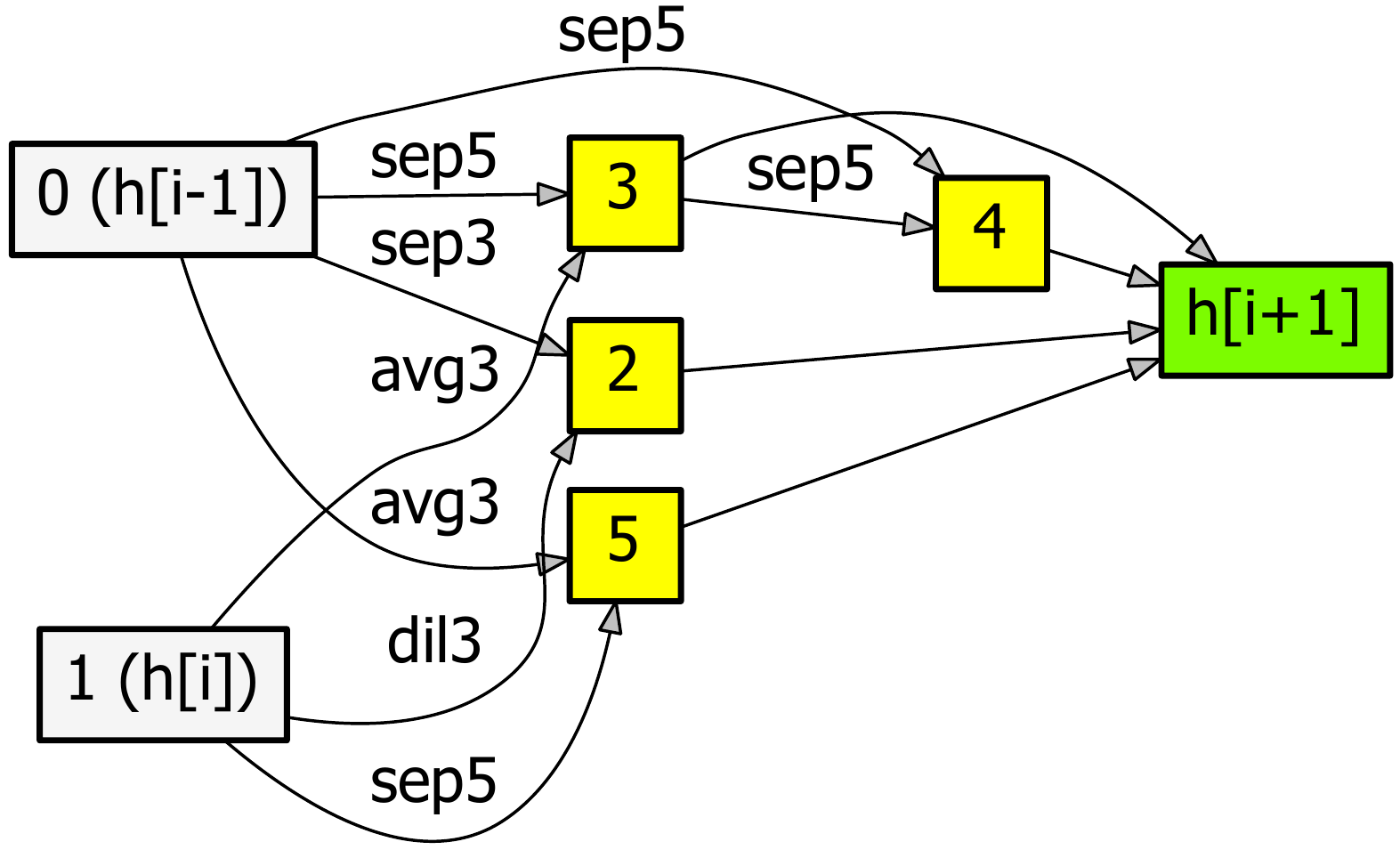}
\vspace{0.2cm}
\centerline{\small (a) Normal cell}
\includegraphics[width=0.85\linewidth]{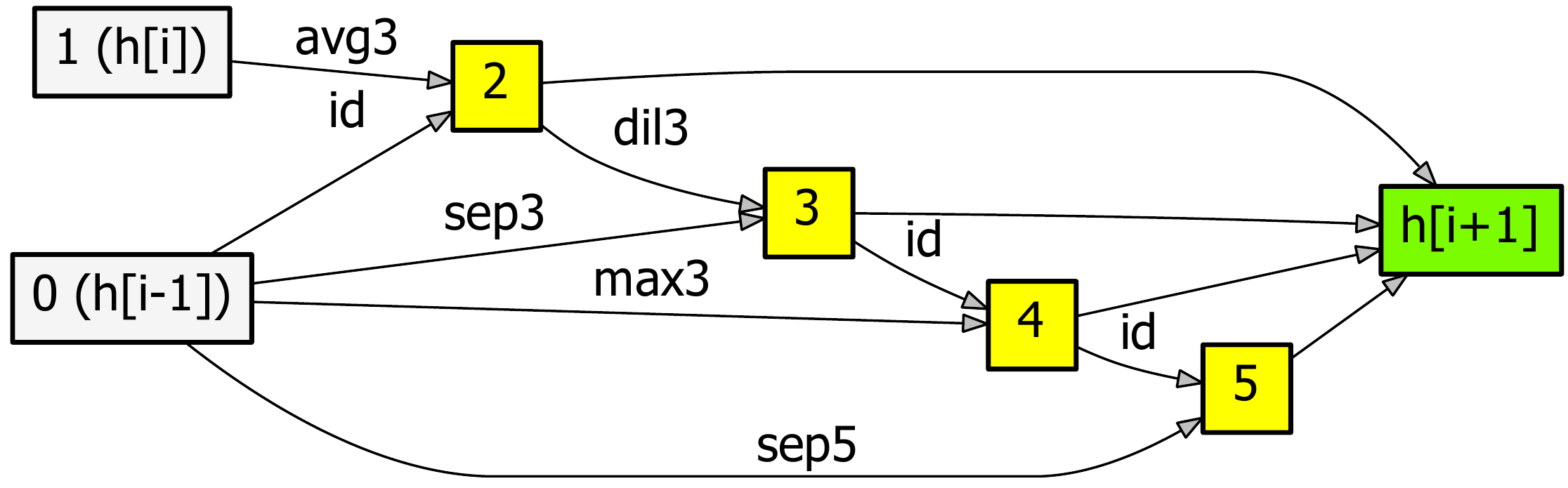}
\centerline{\small (b) Reduction cell}
\vspace{-0.2cm}
\caption{The CNN model discovered by EDNAS on CIFAR-10.}
\vspace{-0.5cm}
\label{arch_cifar}
\end{center}
\end{figure}

\begin{table*}[t]
\caption{Comparison between EDNAS and the state-of-the-art neural architecture search methods for image classification on ImageNet dataset with respect to computational cost, params and test errors.}
\label{imagenet_result}
\vskip 0.05in
\begin{center}
\scalebox{0.85}{
\begin{tabular}{clccccc}
\toprule
& \multicolumn{1}{c}{}& \multicolumn{1}{c}{Search Cost} & \multicolumn{1}{c}{Params} & \multicolumn{2}{c}{Test Error (\%)} & \multicolumn{1}{c}{}\\ \cmidrule{5-6}
Category & \multicolumn{1}{l}{Method} & \multicolumn{1}{c}{(GPU days)} & \multicolumn{1}{c}{(M)}& top-1 & top-5 & Search Method\\

\midrule
\multirow{4}{*}{Manual} & Inception-v1~\cite{szegedy2015going} & - & 6.6 & 30.2 & 10.1 & manual\\
&MobileNet~\cite{howard2017mobilenets} & - & 4.2 & 29.4 & 10.5 & manual\\
&ShuffleNet 2$\times$ (v1)~\cite{zhang2017shuffle} & - & 5 & 29.1 & 10.2 & manual\\
&ShuffleNet 2$\times$ (v2)~\cite{zhang2017shuffle} & - & 5 & 26.3 & - & manual\\
\midrule
\multirow{7}{*}{Transfer learning} & NASNet-A~\cite{zoph2018learning} & - & 5.3 & 26.0 & 8.4 & RL\\
&AmoebaNet-C~\cite{real2018regularized} & - & 6.4 & 24.3 & 7.6 & EA\\
&PNAS~\cite{liu2018progressive} & - & 5.1 & 25.8 & 8.1 & RL\\
&DARTS~\cite{liu2018darts} & - & 4.9 & 26.9 & 9.0 & gradient\\
&GHN~\cite{zhang2018graph} & - & 6.1 & 27.0 & - & hypernet\\
&DSO-NAS~\cite{zhang2018you} & - & 4.7 & 26.2 & 8.6 & gradient \\
\cmidrule{2-7}
&EDNAS & - & 5.2 & 26.8 & 8.9 & RL\\
\midrule
\midrule
\multirow{2}{*}{Directly search} & DSO-NAS~\cite{zhang2018you} & 6 & 4.8 & 25.4 & 8.4 & gradient \\
\cmidrule{2-7}
&EDNAS & 3.67 & 4.7 & 26.9 & 8.9 & RL\\
\bottomrule
\end{tabular}
}
\end{center}
\vskip -0.1in
\end{table*}

To further analyze the search procedure, we present the statistics of the sampled architectures during our training procedure in Figure~\ref{histograms}.
Specifically, we illustrate the cumulative distributions of the sampled edges of the DAG and the operations in the normal and reduction cells over every 50 epoch.
Figure~\ref{edges_stat_normal} shows that $e_{0,3}$ and $e_{1,3}$ are selected more frequently at the later stage of training while the sampling ratio of $e_{2,3}$ drops consistently over time.
In general, the edges from input nodes are preferred to the ones from the hidden nodes.
On the other hand, when we observe the operation sampling patterns in the normal cell, we can see the clear tendency of individual operations; the frequency of pooling (max, avg) and identity (id) operations decreases gradually while the separable convolutions and dilated convolutions with a relatively large kernel size ($5 \times 5$) become more popular at the later stage of the optimization process.
It implies that the searched models attempt to extract the high-level information from the inputs to improve accuracy.
The statistics in the reduction cells do not change much over time for both edge and operation samples.

The derived architectures of the normal and reduction cells are demonstrated in Figure~\ref{arch_cifar}.
The characteristics of the two models are different in the sense that the normal cell has many parallel operations which coincides with the tendency illustrated in Figure~\ref{edges_stat_normal} while the operations in the reduction cell tend to be serial.

\subsection{Convolutional Cell Search with ImageNet}

\subsubsection{Data and Experiment Setting} 
The ImageNet dataset contains almost 1.2M images in 1,000 classes for training~\cite{deng2009large}. For our architecture search, we use 1M images to train the model and 0.2M images to train policy vectors. 
%
Our algorithm for ImageNet has the exactly same search space with CIFAR-10 except that it has the additional stem module to convert input images from $224 \times 224$ to $28 \times 28$, which is the similar size to the input images in CIFAR-10.

We employ the SGD optimizer for architecture search without the Nesterov momentum, where the initial learning rate is 0.05, which is reduced by the factor of 10 at every 10 epoch.
Adam is used for policy vector search with learning rate 0.00035.
The batch sizes for training model and policy vector search are 200 and 50, respectively. 
The training is carried out for 50 epochs.


The architecture for performance evaluation is composed of 14 cells (12 normal cells and 2 reduction cells), where each reduction cells follows a series of 4 normal cells.
Also, we integrate the auxiliary classifier to learn the model using an SGD with the learning rate 0.1, the batch size 200, and the network is trained for 250 epochs.

\begin{figure}[t]
\begin{center}
\includegraphics[width=0.65\linewidth]{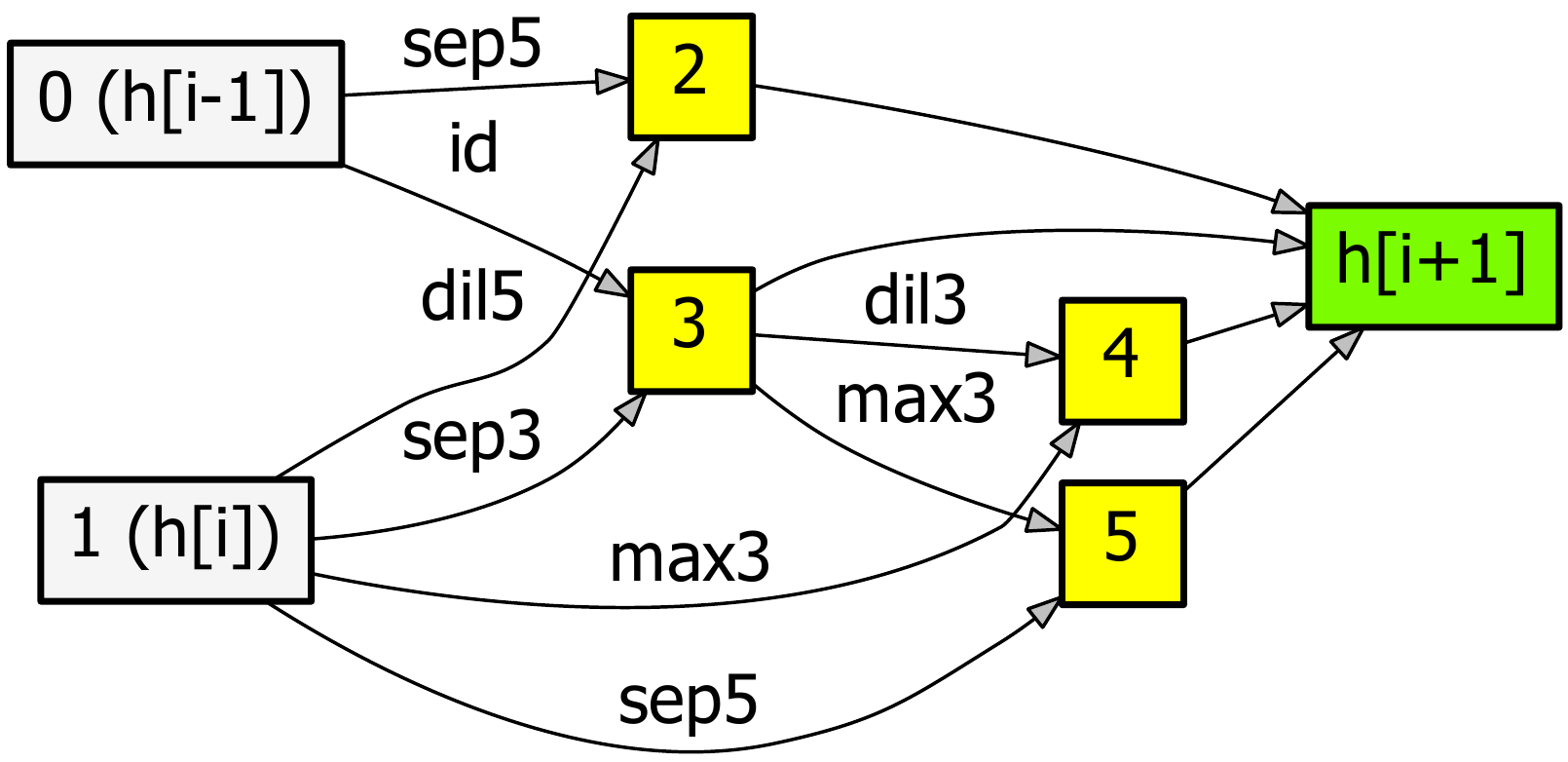}
\vspace{0.2cm}
\centerline{\small (a) Normal cell}
\includegraphics[width=0.65\linewidth]{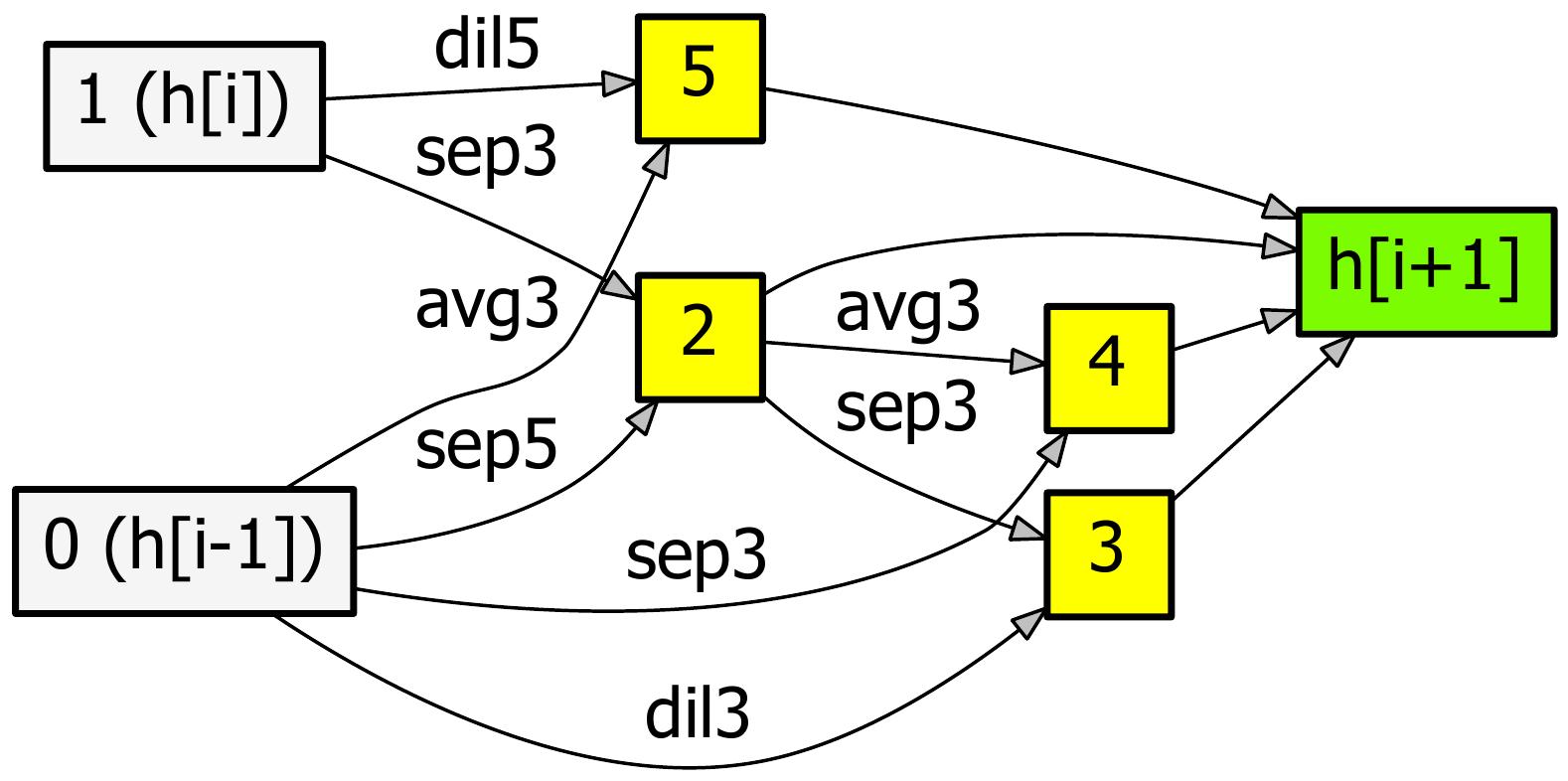}
\centerline{\small (b) Reduction cell}
\vspace{-0.3cm}
\caption{The CNN model discovered by EDNAS on ImageNet.}
\vspace{-0.8cm}
\label{arch_imagenet}
\end{center}
\end{figure}

\begin{table*}[t!]
\caption{Comparison between EDNAS and the state-of-the-art neural architecture search methods for language modeling on Penn Treebank dataset with respect to computational cost, params and test perplexity.}
\label{ptb_result}
\vskip 0.1in
\begin{center}
\scalebox{0.85}{
\begin{tabular}{clcccc}
\toprule
& \multicolumn{1}{c}{}& \multicolumn{1}{c}{Search Cost} & \multicolumn{1}{c}{Params} & \multicolumn{1}{c}{} & \multicolumn{1}{c}{}\\
Category & \multicolumn{1}{l}{Method} & \multicolumn{1}{c}{(GPU days)} & \multicolumn{1}{c}{(M)}& \multicolumn{1}{c}{Test PPL} & Search Method\\
\midrule
\multirow{2}{*}{Manual} & LSTM~\cite{merity2017regularizing} & - & 24 & 58.8 & manual\\
& LSTM + 15 softmax experts~\cite{yang2017breaking} & - & 22 & 56.0 & manual\\
\midrule
\multirow{10}{*}{\shortstack{NAS \\ methods}} & NAS~\cite{zoph2016neural} & $10^4$ CPU days & 25 & 64.0 & RL\\
& ENAS~\cite{pham2018efficient} & 0.5 & 24 & 55.8 & RL \\
& ENAS (reproduction)~\cite{liu2018darts} & 0.5 & 24 & 63.1 & RL \\
& DARTS (first order)~\cite{liu2018darts} & 0.13 & 23 & 60.5 & gradient \\
& DARTS (first order)$^\dagger$~\cite{liu2018darts} & 0.09 & 23 & 64.2 & gradient\\
& DARTS (second order)~\cite{liu2018darts} & 1 & 23 & 56.6 & gradient\\
& NAONet~\cite{luo2018neural} & 300 & 27 & 56.0 & gradient\\
& NAONet-WS~\cite{luo2018neural} & 0.4 & 27 & 56.6 & gradient\\
\cmidrule{2-6}
& Random & 0.10 & 23 & 61.24 & -\\
& EDNAS & 0.11 & 23 & 59.45 & RL\\
\bottomrule
\end{tabular}
}
\end{center}
\vskip -0.1in
\end{table*}

\subsubsection{Results and Discussion} 
Table~\ref{imagenet_result} presents the overall comparison with other methods on ImageNet dataset.
Most of the existing approaches identify the best architecture on CIFAR-10 dataset and use the same model to evaluate performance on ImageNet after fine-tuning.
This is mainly because their search costs are prohibitively high and it is almost impossible to apply their algorithms on the large-scale dataset directly.
Contrary to these methods, DSO-NAS~\cite{zhang2018you} and our algorithm, denoted by EDNAS, are sufficiently fast to explore the search space directly even on the ImageNet dataset.
The performance of EDNAS is as competitive as DSO-NAS in terms of model size and accuracy, but the search cost of EDNAS is substantially smaller than DSO-NAS.

Figure~\ref{arch_imagenet} illustrates the identified normal and reduction cells, which have the same graph topology while operations are somewhat different.


\subsection{Recurrent Cell Search with PTB}

\subsubsection{Data and Experiment Settings}
Penn Treebank dataset is a widely-used benchmark dataset for language modeling task.
Our experiment is conducted on the standard preprocessed version~\cite{zaremba2014recurrent}.
%
For architecture search, we set the embedding size and the hidden state size to 300, and train the child network using the SGD optimizer without momentum for 150 epochs.
We set the learning rate to 20 and the batch size to 256. 
Also, dropout is applied to the output layer with a rate of 0.75.
To train the policy vectors, we use the Adam optimizer with learning rate $3\times10^{-3}$.

For evaluation, the network is trained using the averaged SGD (ASGD) with the batch size 64 and the learning rate 20.
The network is trained for 1600 epochs.
The embedding size are set to 850, and the rest of hyper-parameters are identical to the architecture search step.
\begin{figure}[t]
\begin{center}
\includegraphics[width=\linewidth]{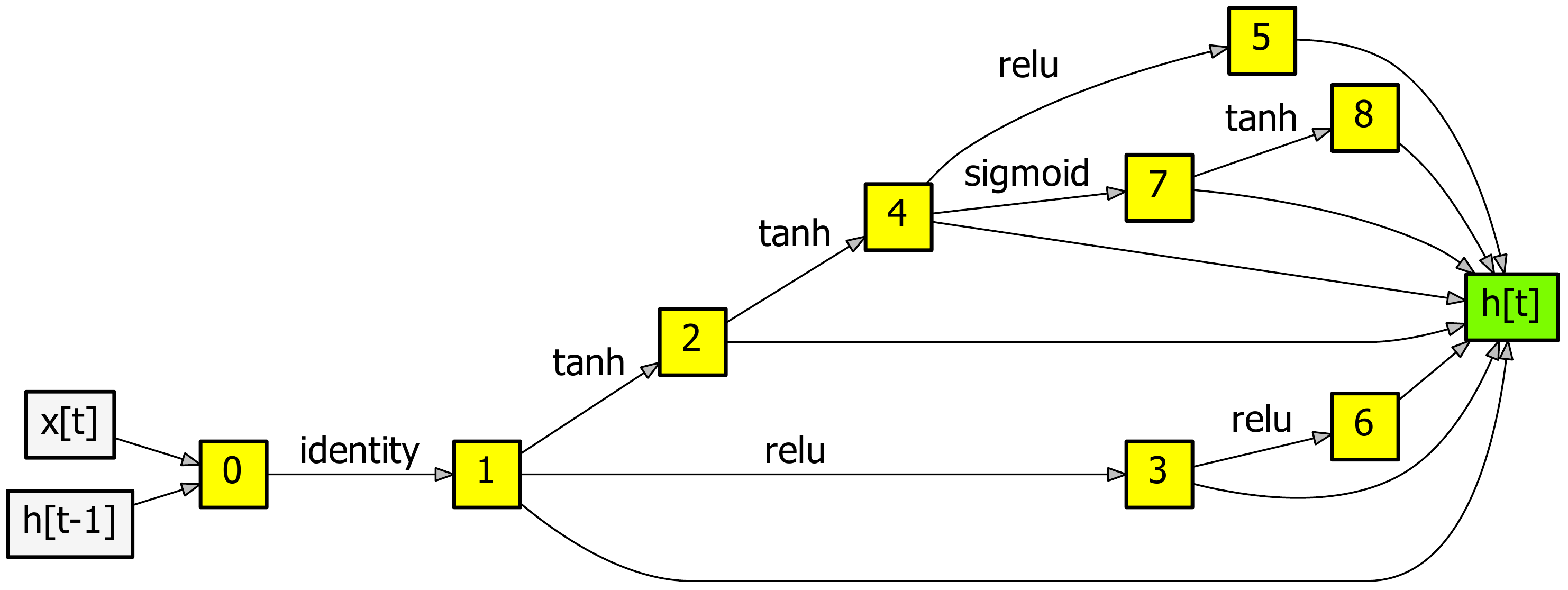}
\vspace{0.1cm}
\caption{The RNN model obtained by EDNAS on Penn Treebank.}
\label{rnn_discovered_cell}
\end{center}
\end{figure}

\vspace{-0.3cm}
\subsubsection{Results and Discussion}
Table~\ref{ptb_result} illustrates the comparison results of various NAS methods in RNNs.
The best architecture discovered by EDNAS achieves 59.45 in terms of test perplexity, and its search cost is 0.11 GPU days only.
The performance of EDNAS is competitive compared to the other NAS methods.
EDNAS shows better accuracy than NAS and DARTS (first order) while it is several times faster than most of the other approaches.
%
Figure~\ref{rnn_discovered_cell} presents the best architecture identified by EDNAS in RNN.


\section{Conclusion}
\label{sec:conclusion}
We presented a novel neural architecture search algorithm, referred to as EDNAS, which decouples the structure search and the operation search by applying the separated policy vectors.
Since the policy vectors in EDNAS are fully observable, we can analyze the architecture search process in EDNAS. 
The experimental results demonstrate that the proposed algorithm is competitive in terms of accuracy and model size on the various benchmark dataset.
In particular, the architecture search procedure in EDNAS is significantly faster than most of the existing techniques.

{\small
\bibliographystyle{ieee}
\bibliography{egbib}
}

\end{document}